\documentclass[conference]{IEEEtran}
\usepackage{xcolor}
\IEEEoverridecommandlockouts

\makeatletter
\newcommand{\linebreakand}{%
  \end{@IEEEauthorhalign}
  \hfill\mbox{}\par
  \mbox{}\hfill\begin{@IEEEauthorhalign}
}
\makeatother
\usepackage{enumitem}

\usepackage{cite}
\usepackage{amsmath,amssymb,amsfonts}
\usepackage{algorithmic}
\usepackage{algorithm}
\usepackage{graphicx}
\usepackage{textcomp}
\usepackage{xcolor}
\usepackage{subcaption}

\def\BibTeX{{\rm B\kern-.05em{\sc i\kern-.025em b}\kern-.08em
    T\kern-.1667em\lower.7ex\hbox{E}\kern-.125emX}}
\begin{document}

\title{Clustering Dynamics for Improved Speed Prediction Deriving from Topographical GPS Registrations\\
\thanks{
}
}

\author{
    \IEEEauthorblockN{Sarah Almeida Carneiro}
    \IEEEauthorblockA{\textit{Univ. Gustave Eiffel, LIGM}\\
                      F-77454 Marne-la-Vallée, France \\
                     0000-0001-7653-8614}
    \and%
    \IEEEauthorblockN{Giovanni Chierchia}
    \IEEEauthorblockA{\textit{Univ. Gustave Eiffel, CNRS, LIGM}\\
                      F-77454 Marne-la-Vallée, France \\
                      Paris, France \\
                      0000-0001-5899-689X}
    \and
    \IEEEauthorblockN{Aurélie Pirayre}
    \IEEEauthorblockA{\textit{Control, Signal and System Dept.} \\
                      \textit{IFP Energies nouvelles}\\
                       Rueil-Malmaison, France \\
                       0000-0003-0112-3689}
    \and%
    \linebreakand 
    \IEEEauthorblockN{Laurent Najman}
    \IEEEauthorblockA{\textit{Univ. Gustave Eiffel, CNRS, LIGM}\\
                      F-77454 Marne-la-Vallée, France \\
                      Paris, France \\
                      0000-0002-6190-0235}
}
\IEEEoverridecommandlockouts

\maketitle

\IEEEpubidadjcol

\begin{abstract}
A persistent challenge in the field of Intelligent Transportation Systems is to extract accurate traffic insights from geographic regions with scarce or no data coverage. 
To this end, we propose solutions for speed prediction using sparse GPS data points and their associated topographical and road design features. 
Our goal is to investigate whether we can use similarities in the terrain and infrastructure to train a machine learning model that can predict speed in regions where we lack transportation data. For this we create a Temporally Orientated Speed Dictionary Centered on Topographically Clustered Roads, which helps us to provide speed correlations to selected feature configurations. Our results show qualitative and quantitative improvement over new and standard regression methods. The presented framework provides a fresh perspective on devising strategies for missing data traffic analysis. \end{abstract}


\section{Introduction}
\label{sec:intro}

Intelligent Transportation Systems (ITS) play a crucial role in optimizing transportation networks, promising safer, more sustainable, and integrated mobility solutions. An important aspect of ITS is the acquisition of transportation data, which covers a diverse array of techniques for different scenarios and requirements. Some examples include fixed loop sensors, embedded vehicle sensors, and GPS-enabled applications.

Fixed loop sensors are embedded in roadways to detect the passage of vehicles and provide real-time traffic flow. These sensors offer high accuracy but are limited to specific locations, making them less versatile for broader regional analyses. Embedded sensors, on the other hand, are placed within vehicles and offer mobility, enabling data collection across a wide geographic range. They are particularly useful for studying traffic patterns, but their effectiveness may vary based on the density of sensor-equipped vehicles. 

GPS-enabled mobile applications have gained prominence as a cost-effective means of gathering speed data, leveraging the ubiquity of smartphones. Unlike traditional fixed sensors, GPS mobile apps offer the flexibility to collect data across vast regions without the need for costly infrastructure installation. However, it's essential to note that GPS-based data acquisition is not without its limitations. Signal disruptions in certain areas, such as tunnels or densely built urban environments, can affect data accuracy. Additionally, the effectiveness of GPS apps relies on user participation, and the representativeness of the data may vary based on the demographics and habits of app users. Privacy concerns and data security issues also need to be addressed when collecting GPS data.

Accurate speed predictions from such data enhance transportation efficiency models, aiding intelligent routing and navigation systems. This allows drivers to choose optimal routes based on real-time traffic and environmental factors, reducing emissions. However, speed prediction faces challenges such as traffic pattern variability, dynamic environmental factors, missing data, and data scale. Traffic pattern variability, influenced by rush hours, special events, or accidents, leads to sudden and unpredictable speed changes. Dynamic environmental factors, like weather and road quality, further complicate predictions. Missing data in traffic datasets add complexity, hindering accuracy. Processing large-scale urban data requires robust methods and efficient algorithms for real-time forecasts.

This paper addresses speed prediction based on the use of GPS-enabled mobile applications and geographic information system (GIS) data, taking into account the uneven distribution of this data across different regions in France. In this study, highly populated areas tend to exhibit denser data coverage, as a higher concentration of smartphone users contributes to more comprehensive and fine-grained data. Conversely, in less densely populated regions, the limited presence of active GPS mobile users results in sparser data availability. Nevertheless, this does not imply that these regions lack traffic, therefore, this spatial heterogeneity in the data density poses significant challenges by not only creating skews the representativeness of the data, but also complicates efforts to extrapolate findings to certain regions. 

In our approach, we design a pipeline to create a Temporally Orientated Speed Dictionary Centered on Topographically Clustered Roads. This framework addresses a new way of determining speed patterns from region similarity and road design options followed by temporal constraints. This is done with the creation of speed grids depending on a certain road cluster. The importance of this work is rooted in its ability to provide individuals with insights into the speed behavior across distinct road design scenarios, relying on a defined trajectory and inputs that can be entirely simulated without the need for a specialized software. 

The paper is organized as follows. In Section~\ref{sec:biblio}, we discuss existing works on the topic of speed prediction and highlight the main differences with our method. In Sections~\ref{sec:probdef} and~\ref{sec:method}, important concepts used in this work are clarified, and the proposed method is explained. In Section~\ref{sec:exp}, we describe the experiments performed and compare the achieved results to other standard methods. Finally, some concluding notes and suggestions for future work are presented in Section~\ref{sec:conc}.

\section{Previous Works}
\label{sec:biblio}

Yuan and Li~\cite{yuan2021survey} identified five key data types in traffic prediction: (i) Spatial data (Sd) with only spatial attributes; (ii) Temporal data (Td) focusing on temporal characteristics; (iii) Spatio-temporal Static data (STSd) involving constant spatial and temporal dimensions; (iv) Spatial Static Temporal Dynamic data (SSTDd) with changing temporal elements and fixed spatial variables; and (v) Spatial Dynamic Temporal Dynamic data (SDTDd) where both spatial and temporal variables change.

In the realm of traffic prediction, as outlined by Yuan and Li~\cite{yuan2021survey}, there are three primary study types: (i) traffic classification, (ii) traffic generation, and (iii) traffic forecasting. Classification involves categorizing given data, while generation addresses the challenge of acquiring trainable traffic data by simulating specific environments. Forecasting, the focus of this discussion, entails predicting traffic variables like flow or speed for activities such as monitoring, estimation, or evaluation.

Statistical traffic speed prediction studies have predominantly employed methods like Historical Average (HA), and Auto Regressive Integrated Moving Average (ARIMA)~\cite{smith1997traffic, billings2006application, guin2006travel}. These traditional approaches, rely on traffic speed over a fixed time period to make predictions. This methodology can involve the use of linear combinations of past observations and their lagged differences. While valuable, due to the model's complexity, these approaches struggled to capture the inherent stochastic nature of traffic, characterized by the complex interplay of numerous other external and dependent factors. 

Machine Learning (ML) models offer enhanced generalization capabilities, learning intricate relationships within data. Yin et al.~\cite{yin2021deep} identify three core classical ML methodologies: (i) Feature-based models, addressing traffic prediction with manually crafted traffic features in regression models; (ii) Gaussian process models, coping with inner characteristics using kernel functions, albeit with higher computational load; and (iii) State space models, assuming observations are generated by Markovian hidden states, enabling natural uncertainty modeling and better capturing the latent structure of spatio-temporal data. However, the limited overall non-linearity of these models makes them sub-optimal for modeling complex and dynamic traffic data most of the time.

Mi et al.~\cite{mi2022dynamic},to tackle the challenges associated with ensemble learning and the static nature of modeling in traffic speed prediction,proposed a three-step framework. The initial step involved constructing the traffic speed forecasting model using a recursive network (SRU) and temporal convolution network (TCN). Subsequently, the neural networks were integrated through the optimization of weight coefficients (MOICA). Finally, the optimal solution from MOICA was selected based on changes in the traffic speed data. The dataset utilized in this study was collected from Changsha, and the features incorporated in the model encompassed traffic speed, traffic volume, weather conditions, congestion level, and holiday information.

Zhang et al.~\cite{zhang2022spatial} directed their efforts toward addressing the prediction of medium- to long-term traffic speed while simultaneously minimizing short-term prediction errors. Their approach involved introducing a temporal attention convolutional network (ATCN) and focusing on graph models. Similarly, they conceptualized the observation devices in the network as vertices and the connections between them as edges. The model takes in a weighted adjacency matrix and a historical step traffic speed matrix as inputs and produces a traffic speed matrix for the prediction time step. The data utilized in their study was sourced from PEMS04, PEMS08, and LOS.

Hu et al.~ \cite{hu2022attention} proposed a framework for the prediction of large-scale traffic speed, employing the AB-ConvLSTM. The model incorporates a convolutional-long short-term memory (Conv-LSTM) module, an attention mechanism module, and two bidirectional LSTM (Bi-LSTM) modules. To gauge its effectiveness, the model underwent a thorough evaluation against ARIMA, RF, SVR, LSTM, SAE, and CNN AT-BLSTM.

Ma et al.~\cite{ma2022novel} proposed a method for predicting traffic speed that relies on incorporating both spatial-temporal information and selecting optimal inputs based on their predictive accuracy on the validation set. The dataset comprised seven months of 2015 data obtained from seven long-distance microwave detectors, monitoring flow, density, and lane occupancy. Following data filtration, their Spatio-Temporal Feature Selection Algorithm (STFSA) identified four monitoring points with the strongest correlation to the predicted points, forming a spatial-temporal correlation matrix. STFSA operates using Pearson’s correlation coefficient for time-series correlation and the Las Vegas method (LVM) for feature selection. This matrix is then employed in a STFSA-CNN-GRU Hybrid Model (SCG) prediction model. A comparative analysis is conducted against established methods, including ARIMA, SVR, CNN, RNN, LSTM, and GRU.

Wang et al.~\cite{wang2023traffic} explored within the context of incomplete data for prediction models, the study addressed the challenges posed by error accumulation resulting from the use of imputation to fill gaps in the data. The authors proposed a Graph Neural Network (GSTAE) trained on spatio-temporal information coupled with an encoder-decoder for parallel speed prediction. The underlying graph structure, denoted as G = (V, E), consisted of a set of traffic sensors (V) and their connectivity (E). Given the difference in objectives of prediction and imputation, the two modules were trained sequentially. This sequential training approach was adopted to leverage imputation in allowing the encoder to derive representations from the initially gapped data. The proposed model's efficacy was evaluated using the PEMSD7 and METR-LA datasets.

Tian et al.~\cite{tian2023mat} introduced a model to predict traffic speed, incorporating a multi-head attention mechanism and a weighted adjacency matrix known as MAT-WGCN. In this approach, a Graph Convolutional Network (GCN) is employed to capture spatial features of the road network within the weighted adjacency matrix. Simultaneously, a Gated Recurrent Unit (GRU) is utilized to extract temporal correlations between speed and time. These spatial and temporal features are then combined and input into a multi-head attention mechanism. The model's performance is evaluated on the EXPY-TKY and METR-LA datasets, and the results are compared with existing methods such as HA, SVR, and various GCN and GRU-based approaches.

Qiu et al.~\cite{qiu2023graph} aimed to enhance predictive accuracy by focusing on event factors. They introduced an event-aware graph attention fusion network (EGAF-Net), incorporating a heterogeneous graph (G = (V, E)) representing road segments and connections. An Event set, including critical attributes like time, location, and speed in surrounding areas, was integrated. Their approach's efficacy was evaluated on the Q-Western-Traffic, Q-Eastern-Traffic, and Q-Traffic datasets.

Within our knowledge and research, works that exhibit a closer resemblance to our proposition are the ones involved in the topic of travel speed prediction (\cite{huang2018sparse,tang2020short,laraki2020vehicle,yang2021fast}). In which Huang et al.~\cite{huang2018sparse} dealt with travel speed prediction with missing traffic information with a PCA based technique and spectral clustering of historical observations of road speed ratios. In their study, Laraki et al~\cite{laraki2020vehicle}, leverage GPS and commercial Geographic Information System (GIS) data to construct a speed profile for a road based on three key points of speeds representing the origin, middle, and destination. By categorizing these three points for each road, they cluster them to define representative driving behaviors for each category, ultimately determining the probable profile of the road. While Yang et al.~\cite{yang2021fast} proposed the prediction of various levels of traveling speed with data acquired form historical trajectory data. They defined in their work a basic path speed cell based on GPS points clustering and were able to predict up until an entire given path with the aid of 1-D convolution layer and bidirectional LSTM in their model. While these studies predominantly center on speed prediction, their primary focus revolves around historical speed data suitable for time-series speed prediction, often sourced from fixed-location sensors. Moreover, there lacks consensus on uniform experimental input data among these methodologies, with only a handful of benchmark datasets applicable to these issues. Consequently, making an informed comparison among them and identifying overarching challenges becomes difficult.

This study builds upon the framework presented in~\cite{carneiro2023swmlp}. However, instead of relying on 3-point feature associations for input into a shared-weight multi-layer perceptron (SWMLP), the current approach adopts a Recurrent Neural Network (RNN) and introduces novel spatio-temporal features. The enhancement lies in diminishing the dependence on fixed-distance point-association, resulting in the development of a more generalized and simulatable scenario. This adjustment continues to tackle the challenge posed by the absence of historical speed data and the complexity associated with adapting models to diverse regions.

\section{Problem Definitions}
\label{sec:probdef}

In this section, we define the problem addressed by our work, that is trajectory point-wise speed prediction for speed profiling in missing data. We also introduce the basic concepts for understanding the problem.

\subsection{Trajectory}
\label{subsec:traj}
We define a trajectory $T$ as a progression of registered GPS points that its latitude and longitude form a parametric curve in Cartesian coordinate system. We can then define a trajectory as $T(n) = \{p_0, p_1, p_2, ...., p_n\}$, in which $p_i = (x,y,f_{p_i})$ is the registered latitude $x$, and longitude $y$ point over the time step $n$. In addition, for each GPS point $p_i$ there is a feature vector $f_{p_i}$ also associated to it.

\subsection{Trip}
\label{subsec:trip}
A trip in this work refers to a journey made using a car involving traveling from one location (origin) to another (destination). Considering $Tp$ a trip, $L$ a road link with and unique identity $id$, and $n$ the number of timesteps in the registered GPS trace, $Tp = \{L^{id_A}_1,L^{id_A}_2,L^{id_B}_3,L^{id_B}_4,L^{id_B}_5, ...,L^{id_G}_{n-1},L^{id_H}_n\}$. This means that a trip is composed of multiple roads links that might or not have distinct ids.

\subsection{Missing Region Information}
\label{subsec:miss}

Based on a mobile application named Geco air\footnote{www.gecoair.fr/en/}, we are able to acquire GPS mobility data in the form of trajectories. Exploratory analysis of this dataset have revealed a heterogeneity in the density of the data acquired across the French territory. Only a small part of the territory is actually represented, mainly the big cities, and in particular Paris. Geco air data collected over approximately 1 year (January 2017 - March 2018) show 5.2 million km. Of these, 30\% of these data, or 1.6 million kilometers, cover only 12 major cities (Paris, Lyon, Marseille, Toulouse, Bordeaux, Lille, Nice, Nantes, Strasbourg, Rennes, Grenoble, Tours).

\subsection{Trajectory Point-wise Speed Prediction}
\label{subsec:pwsp}

Point-wise speed prediction is the process of estimating the probable speed of a moving object, such as a vehicle, at each individual registered point along a given trajectory. This prediction usually involves determining how speed changes as the object moves through space and time. In other words, given a a trajectory $T(n) = \{p_0, p_1, p_2, ...., p_n\}$ for each $p_i$, where $i \in {1, \cdots, n}$, we want to associate to it a probable speed $s_{p_i}$ value based on their feature vector $f_{p_i}$. Therefore we want to forecast for $T(n)$ a speed prediction vector $S(n) = \{s_{p_0}, s_{p_1}, s_{p_2}, ...., s_{p_n}\}$.

\subsection{Road Speed Profiling}

Road speed profiling typically refers to the process of collecting and analyzing data related to vehicle speeds on a particular road or a section of a road. This involves gathering information about the speeds at which vehicles travel on that road under various conditions. Road speed profiling can help identify areas where speed limits may need adjustment, where traffic enforcement measures are required, or where road design improvements are necessary to ensure the safe and efficient flow of traffic. Thus in the case of missing region information (Subsection~\ref{subsec:miss}), it is possible to achieve speed profiling from trajectory point-wise speed prediction (Subsection~\ref{subsec:pwsp}).

\subsection{Point-wise Speed Prediction in Missing Data}
\label{subsec:problemdef}

The absence of registered data at a territorial level, as elaborated in Subsection~\ref{subsec:miss}, presents a significant obstacle to advance research studies to design more optimized solutions concerning mobility and sustainability. Assuming that studies conducted in a single region can universally represent the entire country may result in errors, biases, and challenges when implementing these solutions in real-world scenarios. In response to the challenge of lacking regional data, in this paper we want to verify how can we address trajectory speed prediction based on features that, if not registered can be at least simulated. 

Usually the most common data we can find for a territorial extension is about its road design elements. In this work, we consider road design features such as: length; if it has or not a traffic light in one, none, or both of its directions; the distance a car would be from the point that marks the beginning of the road. In addition, we also take into account the class of the car being used to generate this prediction. These were the chosen elements, since they are easily simulated, even without the need of a software, which are generally time consuming. 

The hypothesis for a point-wise trajectory speed prediction of this paper is supported by the following question: "Can we harness the existing regional data as an inpainting source to predict the probable missing region speed data, leveraging their road design similarities and corresponding temporal conditions?" The issue addressed here is whether it is possible to perform point-wise speed prediction based on a vehicle's trajectory supported by topographical data. Our primary objective is to ascertain whether predictions produce favorable outcomes by grounding the method in easily accessible variables.

\section{Proposed Method}
\label{sec:method}
\begin{figure*}[t]  
  \centering
  \begin{subfigure}{0,4\textwidth}  
    \includegraphics[width=\linewidth]{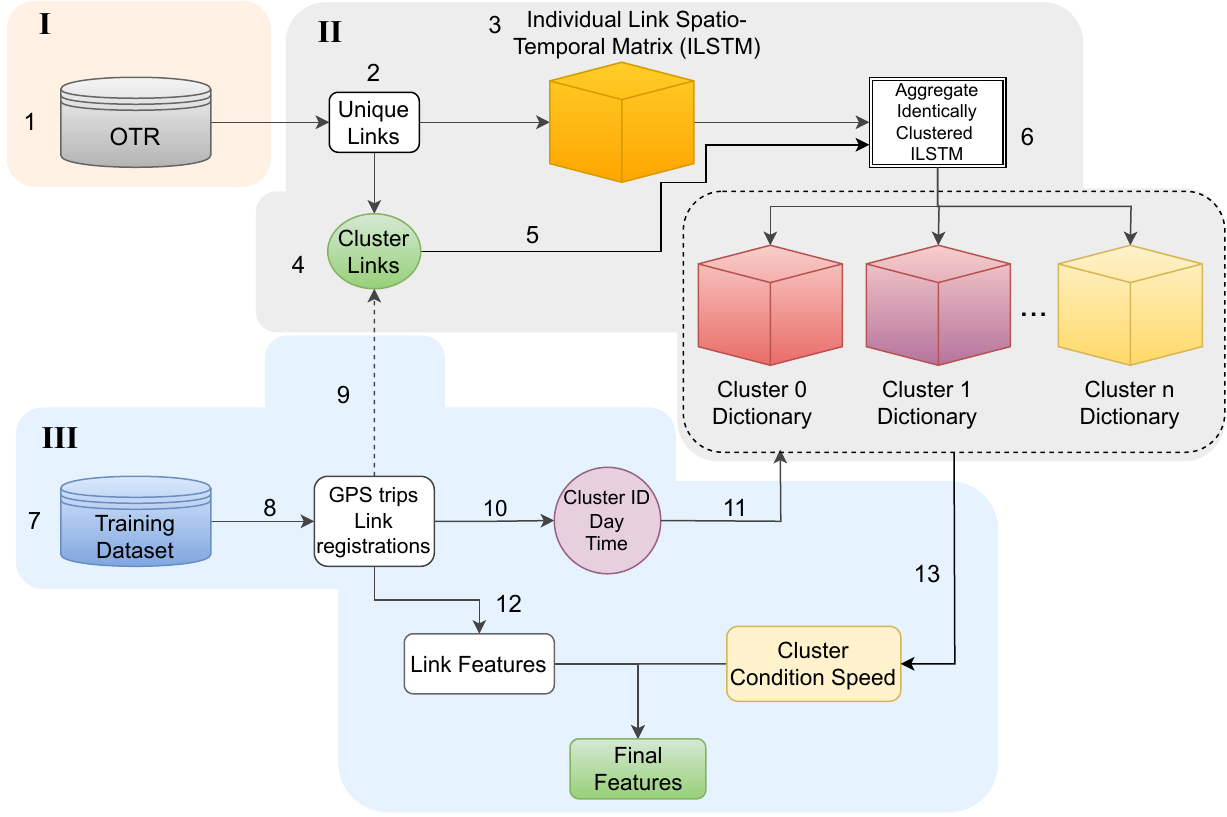}
  \end{subfigure}%
   \hspace{0.02\textwidth}
  \begin{subfigure}{0,35\textwidth}  
    \includegraphics[width=\linewidth]{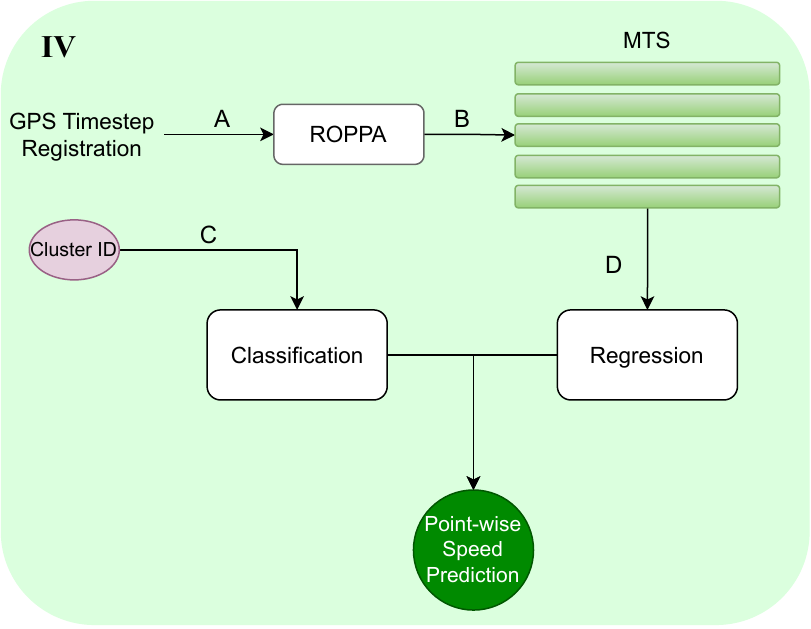}
  \end{subfigure}
  \caption{Diagram of the framework used for this paper. In $I$ (Subsection~\ref{subsec:I})(1) we define an region not used for training the network (OTR) that we have data. In $II$ (Subsection~\ref{subsec:cd}) with the (2) unique links extracted from the OTR we create (3) an Individual Link Spatio-Temporal (ILSTM) for each OTR link. We then (4) cluster the OTR links based on their features and (5) aggregate all of the ILSTM for all of the links that were attributed to the same cluster to create a Cluster Dictionary. Thus, by having all of these cluster dictionaries, we create our (6) Temporally Orientated Speed Dictionary Centered on Topographically Clustered Links. In $III$ (Subsection~\ref{subsec:III}) we work with (7) links that will be used in training, we get the (8) GPS registrations of these links, and we use a (9) link inference clustering process(where each link is assigned to one of the previously calculated clusters). We can then (10) reference the corresponding dictionary (subsection~\ref{subsec:cd}) based on cluster information and temporal details of when the trip occurred. We join the (12) links features to the retrieved (11,13) relevant dictionary feature associated with that specific data point. Finally In $IV$ (Subsection~\ref{subsec:IV}) we classifying an input and associating its loss with the regression process to better train the regression model. }
  \label{fig:method}
\end{figure*}

We propose a method based on a Temporally Orientated Speed Dictionary Centered on Topographically Clustered Links as well as Recurrent Neural Networks (RNN) to predict speed for all given points of a trajectory. In this section, our methodology can be seen in Figure~\ref{fig:method}. The method follows four main sections which are indicated in the diagram by: (I) Off-training region table creation; (II) Cluster dictionaries; (III) Spatio-temporal dictionary value association; and (IV) RNN.

\subsection{Off-training Region Table}
\label{subsec:I}
The first step of this process is to define an off-training region (OTR) in our territory that we have data, which will not be used for training the network. In our case we selected 3 city regions to build these initial tables. This is done so that these regions can serve as the base of inpainting mobility information and do not cause any data in-sample evaluation. The problem of an in-sample evaluation occurs when the model is tested with data it has already seen in training which can lead to misleading impression of the model's performance. Therefore as the idea of this table is to be applied in a variety of regions, it is essential to use a separate dictionary dataset. In our case we selected the regions of Lille, Aix-en-Provence, and Cannes to compose our OTRs.
 
\subsection{Cluster Dictionaries}
\label{subsec:cd}
Here we go through the steps of creating a Temporally Orientated Speed Dictionary Centered on Topographically Clustered Links. The underlying concept of this dictionary involves incorporating its information into the feature vector utilized for speed prediction in instances where speed information for a link is unavailable within a specific temporal scenario.

In this work, we refer to a road section as a link. So we filter all unique links found in all trips inside the OTRs. For each individual link, a matrix is created. This matrix is a $Days \times Hours \times S$ in which $S$ stands for the percentage of the links's subdivision (Algorithm~\ref{algo:speedPat}). The percentage of subdivision is given by the number of divisions (slots) we want to create along a link's length, normalized to percentage, thus, as an example if we want to divide the link information based on 25\%, it means that we will have a total of $S = 4$ slots, and thus link speed information is grouped into placements 0--25\%, 25--50\%, 50--75\%, 75--100\% .

 \begin{algorithm}
\caption{Individual link Spatio-Temporal Matrix}
\begin{algorithmic}[1]
\STATE \# $dist\_list$ where in a link the GPS registration occurred
\STATE \textbf{Input:} ($dist\_list$, $percent$, $ILSTM$, $day$, $hour$,
\STATE                  $link\_speed$)
\STATE
\STATE $slot\_percent\gets$ \textbf{int}(100/percent) 
\STATE \# Initialize
\IF{ there is \textbf{no} content in $ILSTM$}
    \STATE $ILSTM \gets$ []
    \STATE \# Day of the week
    \FOR{$i$ \textbf{in range}($7$)}
        \STATE $hourly\_pattern \gets$ []
        \STATE \# Hour
        \FOR{$j$ \textbf{in range}($24$)} 
            \STATE $slot\_pattern \gets$ []
            \FOR{$k$ \textbf{in range}($slot\_percent$)} 
              \STATE $slot\_pattern$.append([0, []])
            \ENDFOR
            \STATE $hourly\_pattern$.append($slot\_pattern$)
        \ENDFOR
        \STATE $ILSTM$.append($hourly\_pattern$)
    \ENDFOR
\ENDIF
\STATE
\STATE \# Fill in
\FOR{$i$ \textbf{in range} len($dist\_list$)}
    \STATE $speed \gets$ $link\_speed$[i]\# speed at data point
    \STATE$dist \gets$ $dist\_list$[i]
    \STATE$link\_percent\gets$ ($dist$*100)/length

    \STATE$slot \gets$ \textbf{int}($link\_percent$/$percent$) 
    \STATE$ILSTM$[$day$][$hour$][$slot$][0] += 1
    \STATE$ILSTM$[$day$][$hour$][$slot$][1].append(speed)
            
\ENDFOR

\STATE \textbf{Return} $ILSTM$

\end{algorithmic}
\label{algo:speedPat}
\end{algorithm}

In the context of a trip, each link can be associated with multiple trips, occurring at various times and on different days. GPS data records a car position at a given time, allowing us to track its location along a link. With knowledge of the link's length, we can determine if a car is at the initial 25\%, halfway (50\%), or near the end (75\% or 100\%) of the link.

\begin{figure}
    \centering
    \includegraphics[width=0.7\linewidth]{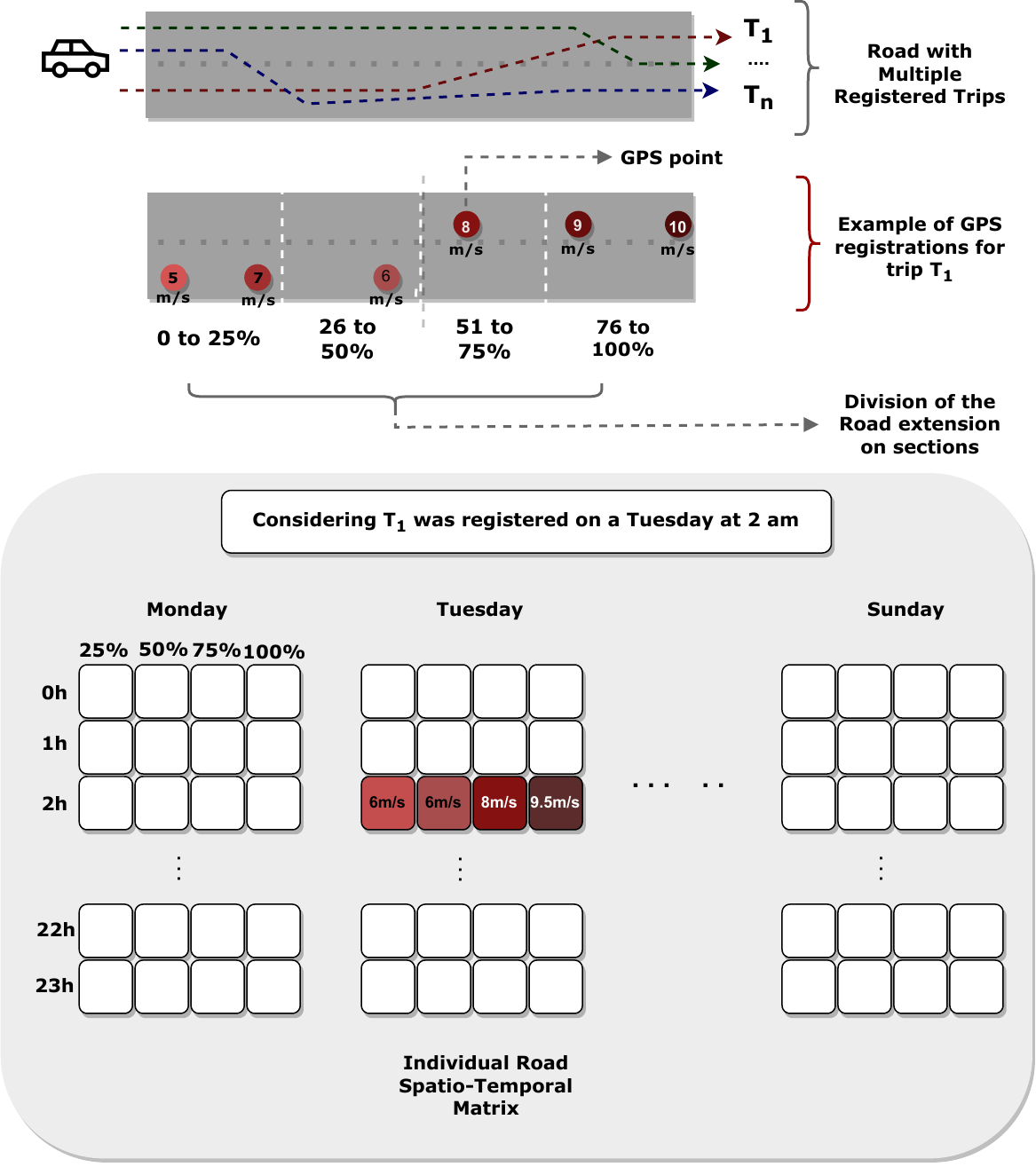}
    \caption{This diagram illustrates the structure of the ILSTM (Individual Link Spatio-Temporal Matrix) and how it is populated. Typically, a link is associated with multiple trips, denoted as $n$. For this explanation, let's assume the link has an identifier $A$, and we will consider $T_1$ as the first trip recorded with $A$ in its GPS data. Consider that $T_1$ occurred at 2 AM on a Tuesday. In this scenario, we can map the corresponding locations within the ILSTM to specific coordinates: depth 1, row 2, and the respective subdivision column. Subsequently, we calculate the average of all data points within each position of the matrix, taking into account the number of data points registered for that particular location. The resulting values are then allocated to the final table. }
    \label{fig:ILSTM}
\end{figure}

Based on these observations, along with the timestamp and day of each registered trip, we can map this information inside the spatio-temporal matrix. After mapping all of the trips for a link, we aggregate data from all trips and calculate average speeds for specific positions within the matrix, taking into account the number of recorded data points at each position. By the end of this process we have a matrix $Days \times Hours \times S$ for all unique id ($L^{id}$) in OTR. This process is illustrated in Figure~\ref{fig:ILSTM}.

All of the OTR have their default topographical information that can be found and calculated based on databases like HERE\footnote{www.here.com}, or Open Street Map (OSM)\footnote{www.openstreetmap.org}. We then select the topographical features: Average, maximal, and minimal link curvature; Average, maximal, and minimal link pitch; Length of the link; link functional class; link speed limit.

\begin{figure}
    \centering
    \includegraphics[width=1\linewidth]{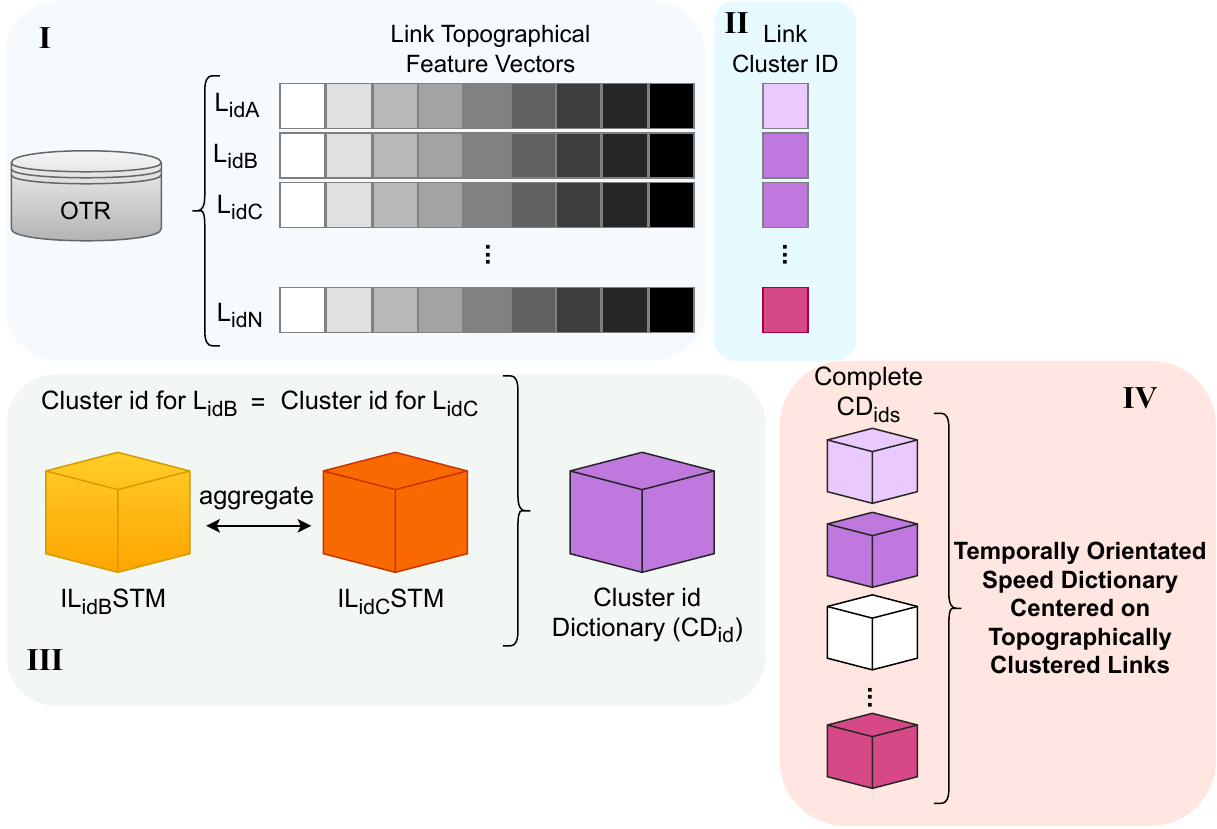}
    \caption{This diagram refers to how we construct a cluster dictionary section (CD) and how their set becomes our Temporally Orientated Speed Dictionary Centered on Topographically Clustered Links. In $I$ the we have the unique OTR links with their topographical features. In $II$ we have the cluster ids for each of these uniques OTR links based on the k-means algorithm. In $III$ all the links that have identical cluster ids have their ILSTM aggregated to from the respective Cluster Dictionary. After all cluster dictionaries are formed, their set ($IV$) is what we call Temporally Orientated Speed Dictionary Centered on Topographically Clustered Links. }
    \label{fig:ClusterDict}
\end{figure}

From these features a K-means clustering algorithm is then used to categorize the links based on their feature similarity. Then all the OTR in the same cluster are averaged (Figure~\ref{fig:ClusterDict}).

\subsection{Spatio-Temporal Value Association}
\label{subsec:III}

Similar to the methodology in Section~\ref{subsec:cd}, the current task involves gathering all links in the training, validation, and testing datasets. After assembling the links and their respective topological features, as previously specified, we apply our trained K-means algorithm. The link features then serve as inputs for an inference clustering process, where each link is assigned to a specific cluster based on the K-means algorithm.

Building upon the trip definition provided in Section~\ref{subsec:trip}, it is understood that each GPS registration corresponds to a particular link. Now, with the identified clusters associated with each link through the inference clustering process, we can reference the corresponding dictionary (subsection~\ref{subsec:cd}) based on cluster information and temporal details of when the trip occurred. This allows us to retrieve the relevant dictionary feature associated with that specific data point.

Subsequent to the identification of features, in this segment of the methodology, we employ two branches of link feature within our database: (i) topographical and (ii) infrastructure design. Drawing from the concept of a trajectory, as discussed in Section~\ref{subsec:traj}, a feature vector ($f_{p_i}$) for all of the gathered points p in a trip $T$ is established. We chose to create two types of feature vectors $f_{p_i}$, each for an independent experimental procedures (further explained in Section~\ref{sec:exp}). For these feature vectors $f_{{p_i}_t}$ and $f_{{p_i}_f}$, the following features have been selected.

\begin{itemize}
\item Topographical features for $f_{{p_i}_t}$:
    \begin{itemize}
        \item Data point curvature ($curv$)
        \item Data point yaw ($yaw$)
        \item Data point elevation ($elv$)
        \item Data point pitch ($ptc$)
        \item Day and hour of the trip ($day, hour$)
    \end{itemize}

\item Infrastructural features for $f_{{p_i}_f}$:
    \begin{itemize}
        \item Link length ($len$)
        \item Position of the car inside the Link ($pos$)
        \item Indication if there is a traffic sign in any of the Link directions ($sign\_start, sign\_stop$)
        \item Car class used ($cc$)
        \item Day and hour of the trip ($day, hour$)
    \end{itemize}
\end{itemize}

The selection of two feature vectors is driven by the recognition that a study can either focus exclusively on fundamental topography or, more commonly, have access to only primary design details concerning the links. For instance, consider scenarios such as link construction, urban planning, and environmental impact assessments. In addition to these features, as we found, in our Temporally Orientated Speed Dictionary Centered on Topographically Clustered Links, the associated cluster dictionary speed ($CDS$) to each of the points of a trip's trajectory we can concatenate it to the corresponding $f_{p_i}$. By the end of this process $f_{p_i}$ can take on the form of either a topographical vector or an infrastructural vector, as in Equations~\ref{featVecT} and~\ref{featVecI}, depending on the experiment.

{\small
\begin{gather}
f_{{p_i}_t} = [curv, yaw, elv, ptc, CDS] \label{featVecT}
\\[0.5em]
f_{{p_i}_f} = [len, pos, sign\_start, sign\_stop, cc, day, hour, CDS]
 \label{featVecI}
\end{gather}
}

\subsubsection{Input Design}

In the final phase, we need to build the inputs to be received by the learning model. For our model we test two distinct input configurations: employing individual points, denoted as $p_i$, as standalone inputs, and adopting micro-time-series (MTS) point associations. Utilizing individual points involves employing the feature vector $f_{p_i}$ as isolated inputs for a model, resulting in predictions for each individual $f_{p_i}$ input. Conversely, point associations refer to a select set of points that constitute a segment of the trajectory to which the given point belongs. In essence, this involves creating micro-time-series for each point within the trip's trajectory to serve as input for the model. 

\begin{figure}
    \centering
    \includegraphics[width=0.71\linewidth]{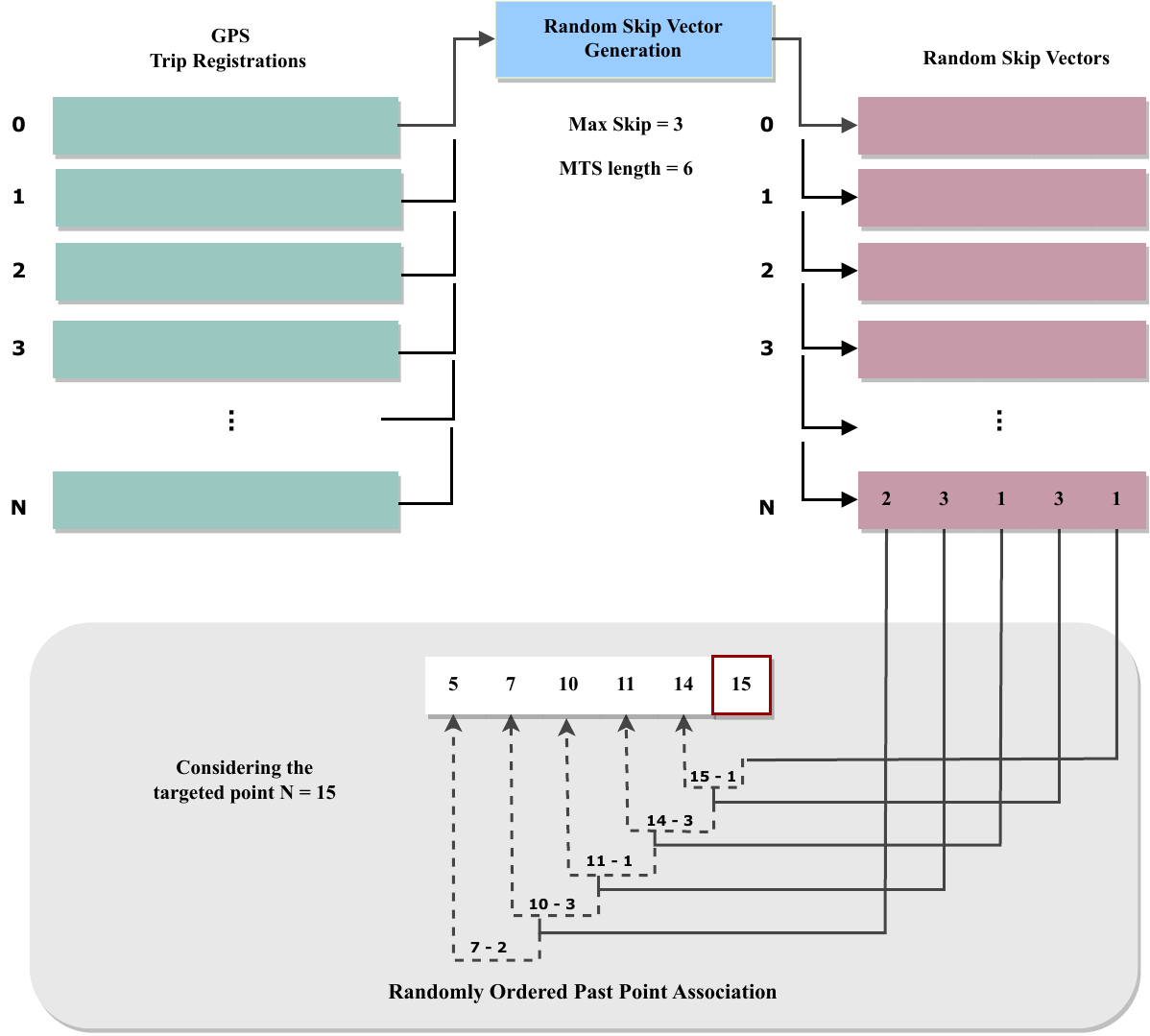}
    \caption{Random Ordered Past Point Association (ROPPA): For each point in a trip trajectory a different random skip vector is generated and becomes a guideline for a ROPPA. As it is possible to see in this example a case in which the point was time step 15, the MTS defined length as equal to 6, and the maximal skip as 3}
    \label{fig:roppa}
\end{figure}

To facilitate the simulation of the MTS in scenarios where precise car positions at each time step are uncertain, we employ a method called Randomly Ordered Past Point Association ROPPA (Figure~\ref{fig:roppa}. For each point $p_i$, we start by defining the MTS size ($Sz$) and creating a random skip array (RSA). The RSA serves as a guide for skipping from one point to another. Since distant time steps may not significantly influence predictions for a given point, our RSA primarily contains small values. To control this randomness, we set a maximal skip distance ($max\_skip$) between consecutive points. Based on this limit, we populate the RSA with $Sz$ random numbers ranging from $1$ to $max\_skip$.

With the RSA in place, we can determine the positions of past points by cumulatively subtracting the RSA values from the current point's position in the trajectory. Importantly, the RSA varies for each input, preventing the model from learning a fixed pattern of time intervals between past points. This adaptability (ROPPA) makes the model better suited for handling non-fixed distance MTS.

ROPPA will be composed of $Sz$ feature vectors $f$ of the points that were associated to the target point $p_n$. Each ROPPA is associated to a label that is the speed registered by the GPS at the time step $n$ creating the possibility to do the point-wise speed prediction and therefore, if the path is complete to create the speed profile for that specific trip trajectory. So for a given time step $p_n$ the ROPPA that is going to be used as input resembles Equation~\ref{roppaFV}. For this equation, the variables $o,v,j,k,q$ are based on the data point indexes calculated by ROPPA's algorithm based on the target index $n$, they must follow the rule that $o<v<j<k<q<n$ and each $f_{p_i}$ refers to Equations~\ref{featVecT} or~\ref{featVecI} for the corresponding data point. If $f_{p_i}$ is chosen according to Equations~\ref{featVecT}, it implies that all corresponding input $f_{p_i}$ instances will also follow Equations~\ref{featVecT}. Similarly, the selection of $f_{p_i}$ based on Equations~\ref{featVecI} results in all associated input $f_{p_i}$ conforming to Equations~\ref{featVecI}.

{\small
\begin{equation}
 f_{{p_n}\_ROPPA} = [f_{p_o},f_{p_v},f_{p_j},f_{p_k},f_{p_q},f_{p_n}]
 \label{roppaFV}
\end{equation}
}

\subsubsection{Model Design}
\label{subsec:IV}

Finally our model is based on two premises a regression and a classification. By classifying an input and associating its loss with the regression process we can help to influence and better train the regression model. 

When we classify an input, we assign it to a specific category based on a predefined criteria. This classification provides the model with a fundamental understanding of the input's nature. Once the input is classified, we can associate the classification loss with the regression process. It is our understanding that by incorporating the classification loss into the regression process, we create a relationship between the two tasks. The regression process aims to predict a continuous or numeric value, often related to the input's characteristics. Therefore, this prediction can be enriched and guided by the knowledge obtained through classification.

From this assumption classification and regression are trained simultaneously. We use the cluster association from the K-means as the category to classify a point, because we are able to assign to what link a point belong to. For the regression aspect, we feed the model with ROPPA inputs (Equation~\ref{roppaFV}). Both sections are trained and their losses weighted and summed by each training iteration.

\section{Experiments}
\label{sec:exp}



We have curated a database using real-world data sourced from IFPEN mobile application Geco Air, and the commercial geographic information systems HERE databases. This database includes car trip data along with the topographical and infrastructure characteristics of the traversed roads, meticulously recorded at each GPS data point.

This database contains a training, validation and testing set. The regions used to create each set were the city of Marseille, Nice and Paris respectively. The selection of these regions was purposeful, driven by their richness in data availability. Although the primary objective of our work is to predict speed profiles for regions lacking data, we conducted these initial experiments within well-documented areas to rigorously validate our methodologies. Furthermore, the training and testing sets were chosen from geographically distant regions to assess the method capacity of generalization.

The training dataset contains around 14000 car trips, while validation and testing 5000 each, with data ranging from the years of 2017 to 2021. The training set contains in its total 2.277.857 registered GPS points, while validation and test have 141.074 and 64.238 respectively.

\subsection{Model}

In the chosen model implementation, an RNN architecture is employed, comprising an RNN layer succeeded by separate branches designed for regression and classification tasks. The regression branch is composed of three fully connected layers (64, 64, and 32 neurons), while the classification branch consists of a fully connected layer followed by a dropout layer of 0.2 and a final fully connected layer (64, 120 neurons). The ReLU activation function is applied after each layer in both branches. The model takes the ROPPA with $CDS$ as input and generates two distinct outputs: one dedicated to regression and another to classification. The classification output exclusively contributes to our loss function, which is formulated as the weighted sum of both branch outputs. The batch size was specified as 1000, the length of ROPPA was set to 5, and the initial learning rate was defined as $10^{-3}$. The loss criterion was the Mean Squared Error, the optimizer employed was Adam, and a Cosine Annealing scheduler was utilized.

\subsection{Metrics and Evaluation}

To evaluate our work we decided to use the metrics that the works found in the literature generally use which are: Mean Squared Error (MSE); Root Mean Square Error (RMSE); and Mean Absolute Error (MAE). The MSE is a commonly used metric in statistics and machine learning to measure the average squared difference between the observed values ($y_i$) and the predicted values ($\hat{y}_i$) in a dataset. The RMSE is closely related MSE, and is particularly more interpretable because it has the same units as the original data, which makes it more intuitive for practical interpretation. The MAE is a metric that calculates the average absolute difference between predicted and actual values in a dataset.

\begin{align*}
   MSE = \frac{1}{n} \sum_{i=1}^{n} (y_i - \hat{y}_i)^2, 
   MAE = \frac{1}{n} \sum_{i=1}^{n} |y_i - \hat{y}_i| \\
\end{align*}
\vspace{-1.8em}
\begin{equation*}
   RMSE = \sqrt{\frac{1}{n} \sum_{i=1}^{n} (y_i - \hat{y}_i)^2}
\end{equation*}

\subsection{Experimental Procedure and Results}

We propose two experimental sets, one utilizing topographical features (i) and the other employing infrastructural features (ii). This decision was made while considering the features that studies may have access for the purpose of speed prediction. Experiments were conducted to verify if our Temporally Orientated Speed Dictionary Centered on Topographically Clustered Roads feature, as well as our use of Random Ordered Past Point Association in an RNN were able to improve trajectory vehicle speed predictions. Our model was compared to standard, and a new regression framework: Multilayer Perceptron (MLP), Random Forest (RF), and the Shared-weight Multi layer perceptron~\cite{carneiro2023swmlp}.

\subsubsection{Predictions with Topographical Features}

In Table~\ref{tab:metricsT} we can observe the quantitative regression analysis, by using the topographical features, for: Multi layer perceptron without using CDS ($MLP\_topo$); Multi layer perceptron using CDS ($MLP\_f\_topo$); and Randomly Ordered Past Point Association Recurrent Neural Network ($ROPPA$ $RNN\_topo$). It is evident that access to data-point-wise topographical information results in relatively low error metrics. This is likely attributed to the non-static values of the feature vectors $f_{p_t}$ utilized. 

\begin{figure}[tb]
    \centering
    \includegraphics[width=1\linewidth]{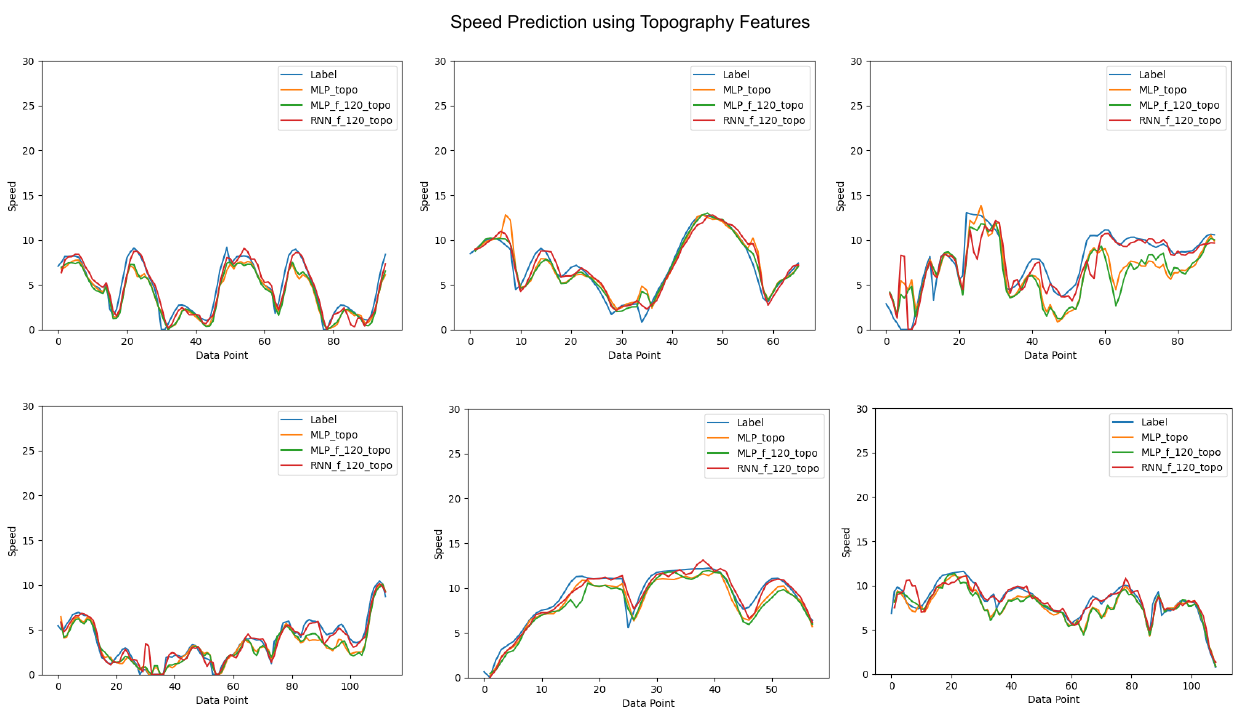}
    \caption{Examples of qualitative speed predictions results using topographical features on 6 different trips of: MLP with no CDS, MLP\_f with CDS, and ROPPA RNN.  }
    \label{fig:PredsTopo}
\end{figure}

Additionally, confirmation of accurate prediction trends can be seen in Figure~\ref{fig:PredsTopo}. It appears that all the networks were able to learn the topographical patterns for predicting speed correctly. Despite the overall good predictions across all networks, the best results are observed when employing the $ROPPA$ $RNN$ in conjunction with $CDS$. However, despite the smoothness of speed prediction achieved with these topographical features, it restricts users from freely inputting arbitrary feature values. Since this is confined to topographical information, it requires the values to be simulated by field experts, becoming less intuitive and less readily available for widespread use. Therefore, in Section~\ref{sub:IF}, we conducted a second set of experiments focusing on features related to the infrastructure of a link, which are more easily simulated and accessible.

\begin{table}[tb]
\centering
\caption{Quantitative analysis of predictions using the topographical features across the entire test set. }
\label{tab:metricsT}
\begin{tabular}{llll}
               & MSE        & RMSE  & MAE\\ \hline
MLP\_topo            & 2.19      & 1.32  & 0.88 \\ \hline
MLP\_f\_topo         & 2.26      & 1.35  & 0.92 \\ \hline
ROPPA RNN\_topo      & 1.16       & 0.97  & 0.59
\end{tabular}
\end{table}

\subsubsection{Predictions with Infrastructure Features}
\label{sub:IF}

In this second set of experiments we now use $f_{p_i}$ as the individual data point feature vectors. Here we anticipated that regression becomes much more challenging as the values of the feature vector components become less dynamic.

We can observe in Table~\ref{tab:metrics} a quantitative analysis made upon the regression results yielded by: Random Forest without using CDS ($RF$); Random Forest using CDS ($RF\_f$); Multi layer perceptron without using CDS ($MLP$); Multi layer perceptron using CDS ($MLP\_f$); Shared-weight Multi layer perceptron with Punctual Past point association arrangement using CDS ($SWMLP$ $PuP$); and Randomly Ordered Past Point Association Recurrent Neural Network ($ROPPA$ $RNN$). Although these models may not achieve the top performance in this comparison, both $RF$ and $MLP$ demonstrate the benefits of incorporating $CDS$ in the training features for speed prediction. This is evident in the improved error metrics observed in $RF_f$ and $MLP_f$. 

In our study, we have chosen $SWMLP$ $PuP$ from the work of Carneiro et al.~\cite{carneiro2023swmlp} as the direct point of comparison for $ROPPA$ $RNN$. This selection was made because $SWMLP$ $PuP$ closely aligns with the methodology of $ROPPA$ $RNN$ in that the associated past data points are gathered based on a distance $d$ from the target, rather than being immediately previous to the target data point whose speed we intend to predict. 

In contrast, $ROPPA$ $RNN$, by employing randomly ordered past point associations, removes the need for precise vehicle positioning and eliminates reliance on a simulator to accurately determine the likely positioning of the vehicle before the target point. In our implementation of the $ROPPA$ method, we employ a distance of up to 3 jumps in the random skip vector, thus, we set the $SWMLP$ $PuP$ distance $d$ the same, and the point association as specified in the paper (3 points, including the target).

Notably, in Table~\ref{tab:metrics}, $ROPPA$ $RNN$ achieves better results compared to $SWMLP$ $PuP$ without requiring a same fixed positioning for all inputs along the path. This attribute renders $ROPPA$ $RNN$ more path-independent, offering a pragmatic advantage in scenarios where precise positioning information is unavailable or challenging to ascertain.

\begin{table}[tb]
\centering
\caption{Quantitative analysis of the regression method predictions across the entire test set. }
\label{tab:metrics}
\begin{tabular}{llll}
               & MSE        & RMSE  & MAE\\ \hline
RF             & 33.20      & 5.42  & 4.66 \\ \hline 
RF\_f          &28.17       & 4.98  & 4.32 \\ \hline 
MLP            & 31.03      & 5.10  & 4.48 \\ \hline
MLP\_f         & 22.10      & 4.35  & 3.79 \\ \hline
SWMLP ($PuP$)~\cite{carneiro2023swmlp} & 8.58       & 2.79  & 2.13 \\ \hline
ROPPA RNN      & 6.75       & 2.47  & 1.88

\end{tabular}
\end{table}

\begin{figure}[tb]
    \centering
    \includegraphics[width=1\linewidth]{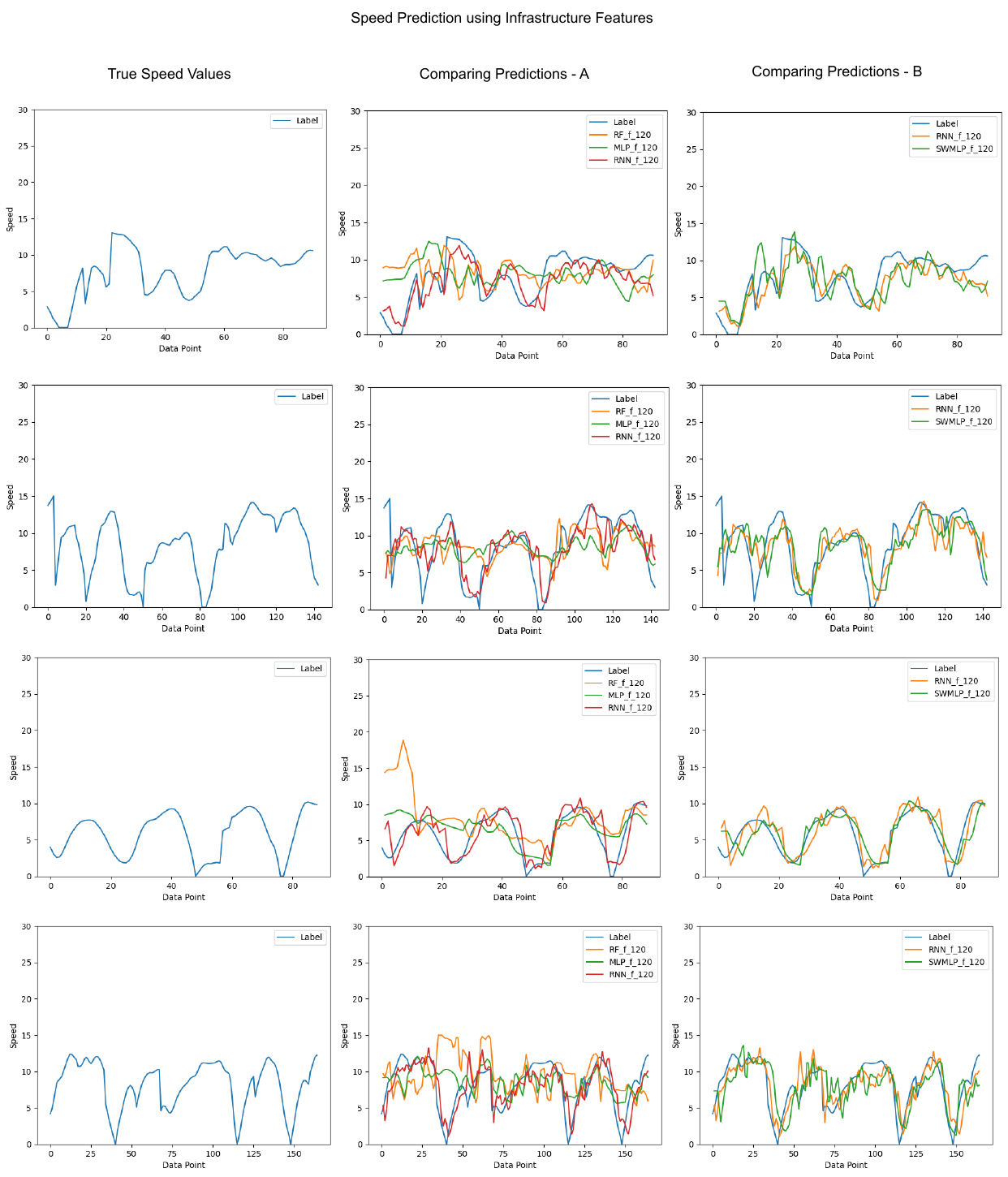}
    \caption{Qualitative comparison of the methods that had the best RMSE values. In this image we have five trips, one in each row, in which the left column represent the true labels (original speed values) registered in each data point of the trip, and middle column the speed regression of: MLP, Random Forest, and our ROPPA RNN. Finally, the right column is the comparison of our method using ROPPA RNN and the SWMLP with the punctual past approach.}
    \label{fig:Prediction-Final}
\end{figure}

For qualitative results (Figure~\ref{fig:Prediction-Final}) we can observe the changes in predictions for the best configurations found in Table~\ref{tab:metrics}: RF\_f ($RF\_f\_120$); MLP\_f ($MLP\_f\_120$); SWMLP PuP ($SWMLP\_f\_120$); and ROPPA RNN ($RNN\_f\_120$). Each row in Figure~\ref{fig:Prediction-Final} corresponds to a distinct trip, where the left column (\textit{True Speed Values}) represents the true labels (original speed values) registered at each data point of the trip. The middle column (\textit{Comparing Predictions-A}) illustrates how the methods (MLP\_f, RF\_f, and our ROPPA RNN) were able to regress the speed values. Finally, the right column (\textit{Comparing Predictions-B}) provides a comparison between our method using ROPPA RNN and SWMLP PuP (the punctual past approach).

In \textit{Comparing Predictions-A}, it is evident that RF\_f and MLP\_f primarily focus on identifying the mean speed value of the data points and struggle to capture speed trends. On the other hand, in \textit{Comparing Predictions-B}, it is noticeable that both SWMLP PuP and ROPPA RNN excel in capturing speed trends in a more explicit manner. 

\subsubsection{OTR Clustering analysis}
\label{sub:OTRClust}

In our experimental configuration, to best define a number of OTR link clusters to use to obtain our presented results we ran an experiment involving 600.000 of the GPS points and defined the OTR clusters number as 6, 30, 60, and 120. In Table~\ref{tab:clusterSens} we can verify that the evaluation metrics suggest that a higher number of clusters can generate less prediction error. Thus, we used 120 OTR link clusters, and we opted for a road division of 10\% for the development of our dictionary. The idea behind increasing the number of clusters was grounded in the potential for generating more specialized groups, as opposed to a generalized cluster. Therefore, enhancing significance through aggregation reduction.

\begin{table}[tb]
\centering
\caption{OTR cluster sensitivity analysis: Metrics found for insightful OTR cluster attribution during experiments. }
\label{tab:clusterSens}
\begin{tabular}{llll}
 OTR Clusters              & MSE        & RMSE  & MAE\\ \hline
6           & 14.73      & 3.36  & 2.80 \\ \hline
30           & 12.53      & 3.18  & 2.68 \\ \hline
60         & 11.64      & 3.06  & 2.58 \\ \hline
120      & 10.96       & 2.92  &  2.45
\end{tabular}
\end{table}

\section{Conclusion}
\label{sec:conc}

In this paper, we present a speed prediction approach utilizing the Temporally Orientated Speed Dictionary Centered on Topographically Clustered Roads, alongside the incorporation of Random Ordered Past Point Association in an RNN (ROPPA RNN) using two different feature approaches. The motivation for this study stems from trajectory speed prediction, in which point-wise topographical elements and road infrastructure design can be considered for predicting point-wise speed.

The key advantages of our approach lie in its capability to leverage similar topographical points with known speed associations to create and map probable speeds based on the temporal context of an input. Through our experiments, we validate that the feature (CDS) generated from the Temporally Orientated Speed Dictionary Centered on Topographically Clustered Roads, combined with Random Ordered Past Point Association, effectively trains a robust RNN across distinct territorial regions (north vs. south).

Furthermore, our findings demonstrate that the enhancements introduced by ROPPA RNN and CDS enable the model to receive inputs with reduced dependence on precise location information. This not only diminishes reliance on an actual simulator for precise data point associations but also provides the option to use more accessible link features while predicting speed, rendering the user more independence in the modeling process.

\bibliographystyle{IEEEbib}
\bibliography{refs}

\end{document}